\documentclass[conference]{IEEEtran}
\IEEEoverridecommandlockouts

\usepackage{cite}
\usepackage{amsmath,amssymb,amsfonts}
\usepackage{algorithmic}
\usepackage{graphicx}
\usepackage{textcomp}
\usepackage{xcolor}
\usepackage{algorithm}
\usepackage{booktabs}
\usepackage{multirow}
\usepackage{array}
\usepackage[table]{xcolor}
\usepackage{float}
\usepackage{url}
\def\BibTeX{{\rm B\kern-.05em{\sc i\kern-.025em b}\kern-.08em
    T\kern-.1667em\lower.7ex\hbox{E}\kern-.125emX}}
\begin{document}

\title{Reasoning in Diffusion Large Language Models is Concentrated in Dynamic Confusion Zones}

\author{\IEEEauthorblockN{Ranfei Chen\textsuperscript{\dag}}
\IEEEauthorblockA{\textit{Institute of Computing Technology} \\
\textit{Chinese Academy of Sciences}\\
Beijing, China \\
chenranfei22@mails.ucas.ac.cn}
\and
\IEEEauthorblockN{Ming Chen\textsuperscript{\dag}}
\IEEEauthorblockA{\textit{Computer Network Information Center} \\
\textit{Chinese Academy of Sciences}\\Beijing, China \\
chenming24@mails.ucas.ac.cn}
\and
\IEEEauthorblockN{Kaifei Wang}
\IEEEauthorblockA{\textit{Institute of Computing Technology} \\
\textit{Chinese Academy of Sciences}\\Beijing, China \\
wangkaifei20@mails.ucas.ac.cn}
\\
\thanks{\textsuperscript{\dag}These authors contributed equally to this work.}
}

\maketitle

\begin{abstract}
Diffusion Large Language Models (dLLMs) are rapidly emerging alongside autoregressive models as a powerful paradigm for complex reasoning, with reinforcement learning increasingly used for downstream alignment. Existing trajectory-based RL methods uniformly allocate policy gradients across denoising steps, implicitly treating all steps as equally important. We challenge this assumption by analyzing trajectories with several step-level metrics: entropy-based uncertainty, Confidence-Margin (CM) uncertainty, and Rate of Entropy Change (RoEC). These reveal structured “zones of confusion”: transient spikes in uncertainty and instability that strongly predict final success or failure, while most steps remain stable. We propose Adaptive Trajectory Policy Optimization (ATPO), a lightweight step-selection strategy that dynamically reallocates gradient updates to these high-leverage steps without changing the RL objective, rewards, or compute budget. Using a hybrid RoEC+CM rule, ATPO delivers substantial gains in reasoning accuracy and training stability across benchmarks, showing that exploiting trajectory dynamics is key to advancing dLLM RL.\footnote{Our code is publicly available at: \url{https://github.com/Aczy156/ATPO}.}
\end{abstract}


\section{Introduction}
\label{sec:intro}
The landscape of large-scale language modeling is rapidly evolving, with diffusion Large Language Models (dLLMs) \cite{nie2025large, ye2025dream, song2025seed} developing alongside traditional autoregressive (AR) models and gaining increasing attention as an alternative generative paradigm for complex reasoning tasks \cite{gong2025diffucoder, zheng2025continuously}. To unlock their full potential and align them with human objectives, reinforcement learning (RL) is indispensable. Recent work has made RL on dLLMs practically feasible through trajectory-consistent training schemes \cite{he2025mdpo, yang2025taming, wang2025revolutionizing}, variance-reduced estimators and stabilized objectives \cite{rojas2025improving, zhu2025llada1_5, wang2025spg, xue2025advantage, tang2025wd1}, and test-time scaling and decoding techniques \cite{chen2025rfg, wu2025fast, liu2025dllm, song2025sparse, wu2025fastv2, wang2025creditdecoding, ma2025dinfer}. 

Despite this progress, large-scale RL training for dLLMs remains brittle in practice: runs often exhibit high gradient variance, unstable learning curves, and extreme sensitivity to the choice of checkpoint. We argue that a central, but rarely scrutinized, design choice underlies these symptoms: how the policy gradient budget is allocated along the trajectory.

Current trajectory-based methods \cite{he2025mdpo, yang2025taming, wang2025revolutionizing, wang2025d2} almost universally adopt a pragmatic shortcut to remain computationally tractable: \textbf{uniform subsampling}, where policy gradients are estimated from only a sparse, evenly spaced subset of trajectory steps. This design choice is built on a powerful but largely unexamined assumption: \textbf{all steps in the reasoning process are of equal importance}. In other words, most existing RL methods for dLLMs implicitly assume that every denoising step is equally valuable for learning, thus it suffices to estimate gradients from a uniformly subsampled subset of the trajectory.

\textbf{This paper argues that this assumption is not just an oversimplification—it is flawed and a primary bottleneck hindering progress.} By treating a highly dynamic reasoning process as a uniform sequence, existing methods waste valuable computation on trivial steps of routine elaboration while failing to capture the brief, decisive moments where model's reasoning is forged. This leads to high-variance gradients, unstable training, and ultimately, suboptimal performance.

Our contribution is to replace this flawed assumption with a data-driven understanding of the model's internal reasoning dynamics. We conduct a systematic analysis of dLLM denoising trajectories and reveal a highly non-uniform landscape of reasoning. We discover two key properties: 1) trajectories are punctuated by identifiable \textbf{``zones of confusion''}—periods of high uncertainty and abrupt belief changes that are highly predictive of the final outcome; and 2) the locations of these critical junctures are \textbf{dynamic}, shifting throughout training and across different problems. These findings motivate a simple question: \emph{if only a handful of steps truly matter, why should we treat all of them the same in policy optimization?}

In summary, this paper makes three contributions:
\begin{itemize}
    \item \textbf{Analysis of trajectory dynamics.} We provide, to our knowledge, the first systematic analysis of reasoning dynamics in dLLM denoising trajectories using entropy, confidence margin, and Rate of Entropy Change. We show that trajectories exhibit highly structured and dynamically evolving \emph{zones of confusion} whose locations shift over training and are strongly correlated with final success or failure, thereby invalidating the common assumption of uniformly important steps.
    \item \textbf{Principle of adaptive gradient allocation.} Motivated by this analysis, we propose a simple design principle for RL with dLLMs: the policy gradient budget should be allocated adaptively according to trajectory uncertainty and instability, rather than uniformly across steps.
    \item \textbf{ATPO framework and empirical validation.} We instantiate this idea in \textbf{ATPO}, a lightweight framework that replaces uniform subsampling with a hybrid adaptive step-selection strategy based on RoEC and CM, while keeping the RL objective, reward design, and overall compute budget unchanged. ATPO is compatible with existing trajectory-based RL methods such as d2, GDPO, and MDPO, and in our experiments it yields consistent improvements in both final accuracy and training stability on several challenging reasoning benchmarks.
\end{itemize}

Taken together, reasoning in diffusion language models is concentrated in a few dynamic confusion zones, and reallocating the gradient budget toward these steps is both effective and minimally intrusive. In the remainder of this paper, we first review background and related work, then analyze trajectory dynamics, introduce the ATPO framework, and finally present experiments and discussion.

\section{The Dynamics of Reasoning in Denoising Trajectories}
\label{sec:motivation}

Trajectory-based RL methods typically rely on uniform subsampling \cite{wang2025d2}, estimating policy gradients from a sparse, evenly spaced subset of denoising steps and implicitly assuming that all selected steps contribute equally. Formally, the final advantage $A^{\pi_{\theta}}(y)$ is broadcast uniformly to every selected segment's gradient, regardless of its content or position:
\begin{equation}
    \nabla_{\theta} J(\theta) = \mathbb{E}_{y \sim \pi_{\theta}} \left[ \sum_{i \in \mathcal{S}_{\text{uniform}}} \left( \nabla_{\theta} \log \pi_{\theta}(y_{\text{seg}_i}) \cdot A^{\pi_{\theta}}(y) \right) \right]
\end{equation}
This section challenges this assumption of uniformity, first by formalizing how to quantify trajectory instability, and second by empirically demonstrating that these instabilities are both critical and dynamic.

\subsection{Formalizing Uncertainty and Instability}
To analyze trajectories, we define step-level difficulty signals: scalar quantities computed at denoising step $t$ from the predictive distribution $p_t(y|x)$ that reflect how ``hard'' that step is. We focus on two families of such signals---\emph{static uncertainty metrics} and a \emph{dynamic instability metric}---and compute them per completion token before averaging over completion tokens and over the batch, yielding a batch-level trajectory of difficulty scores indexed by $t$.

\textbf{Static Uncertainty Metrics.} At any single step $t$, we measure the model's uncertainty using two static step-level signals, both computed from the completion-token logits and then averaged over tokens and batch:
\begin{itemize}
    \item \textbf{Entropy-based Uncertainty}: The Shannon entropy $H(p_t) = -\sum_v p_t(v) \log p_t(v)$ measures the "spread" of the predictive distribution. High entropy indicates the model is uncertain, assigning comparable probabilities to many different tokens.
    \item \textbf{Confidence-Margin Uncertainty}: Denoted CM, defined as the difference between the probabilities of the two most likely tokens, $p_{t,1} - p_{t,2}$. A small margin indicates the model is "conflicted" between its top choices, signaling a point of high local uncertainty. In our implementation, we use the \emph{inverse} confidence margin, $1 / (p_{t,1} - p_{t,2})$, as a numerically stable monotone proxy, but conceptually CM still refers to the underlying margin.
\end{itemize}

\textbf{Dynamic Instability Metric.} Beyond static uncertainty, we also track the "instability" or "surprise" between consecutive steps. A stable trajectory should exhibit a smooth, monotonic evolution of the model's beliefs, while instability corresponds to abrupt shifts. We formalize this as the divergence between consecutive predictive distributions, $p_{t-1}$ and $p_t$. A principled measure for this is the Kullback-Leibler (KL) divergence:
\begin{equation}
    D_{KL}(p_t || p_{t-1}) = \sum_v p_t(v) \log \frac{p_t(v)}{p_{t-1}(v)}
\end{equation}
A high KL divergence signifies a sharp change in the model's belief state. However, computing token-level KL divergence at every step is computationally expensive. As a more efficient proxy, we consider the change in a summary statistic of the distribution—its entropy. We define \textbf{RoEC}, the Rate of Entropy Change, as our dynamic step-level difficulty signal, computed on the batch-averaged entropy trace:
\begin{equation}
\text{RoEC}_t = |H(p_t) - H(p_{t-1})|
\end{equation}
A high RoEC value signals that the distribution's shape has changed significantly, making it a computationally feasible and effective heuristic for detecting these "surprise points" or moments of high dynamic instability.

\begin{figure}[H]
    \centering
    \includegraphics[width=0.9\columnwidth]{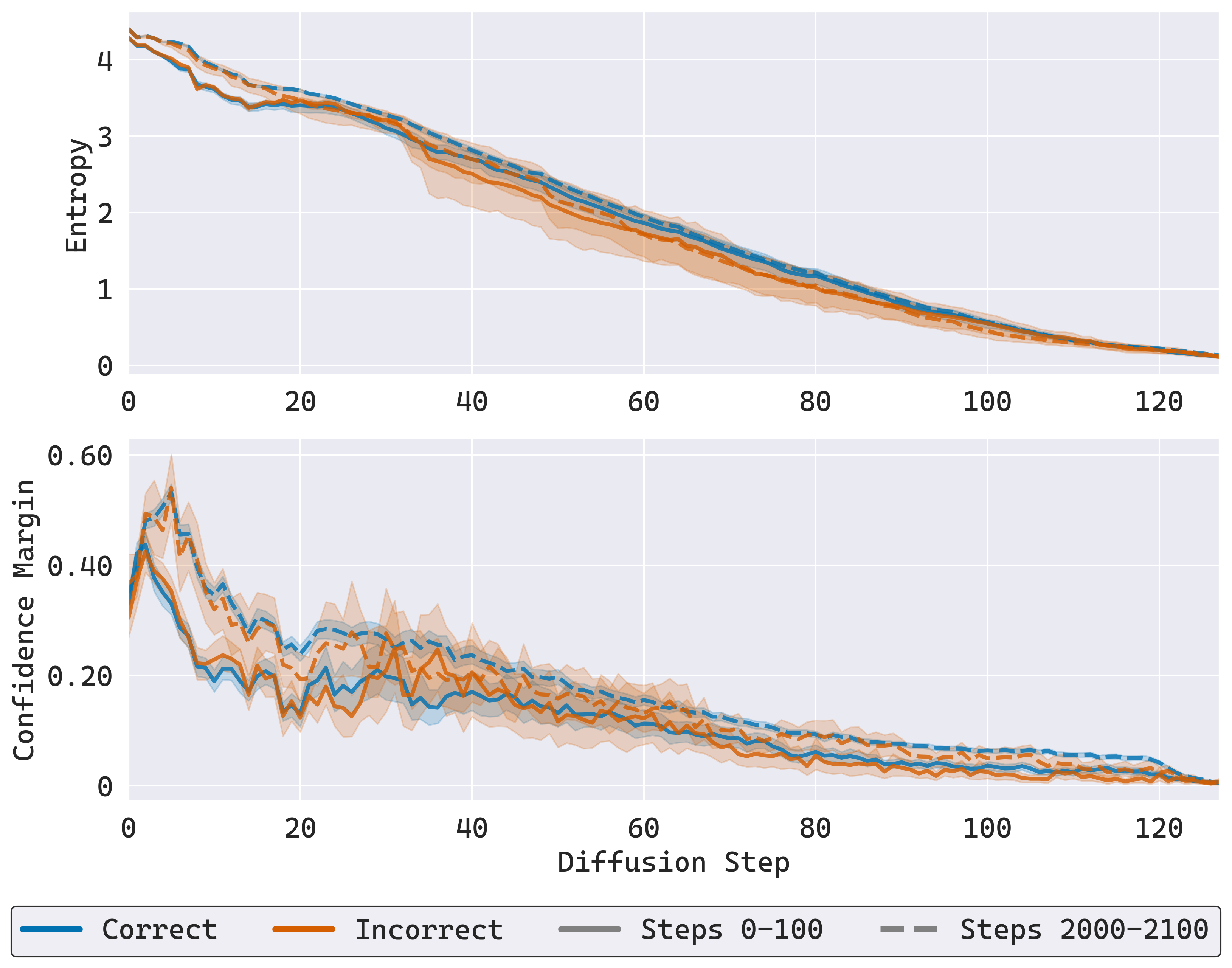}
    \caption{Static uncertainty metrics during denoising on GSM8K. The \textbf{top}
    subplot shows the average Entropy-based Uncertainty, and the \textbf{bottom}
    subplot shows the average Confidence Margin (CM), both plotted over diffusion
    steps. \textbf{Color} distinguishes correct (\textcolor{blue}{blue}) vs.\
    incorrect (\textcolor{orange}{orange}) samples, and \textbf{line style} marks
    an early phase of training (after 100 updates; solid) vs. a late phase of training (after 2000 updates; dashed).
    These results correspond to the ``Static Uncertainty Metrics'' analysis in the
    main text.}
    \label{fig:motivation_dynamics}
\end{figure}

\subsection{Empirical Validation: Signatures of Dynamic Difficulty}

To ground this formalism and connect it to real tasks, we investigate whether these metrics can signal critical junctures in real-world reasoning benchmarks. We analyze the denoising trajectories on the GSM8K dataset, visualizing the average Entropy-based Uncertainty and CM in Figure \ref{fig:motivation_dynamics}. The figure reveals two key insights:

\textbf{1. Uncertainty as an Error Proxy.} First, we test if uncertainty correlates with errors. In Figure \ref{fig:motivation_dynamics}, the curves for incorrect solutions are significantly higher and more volatile than the smooth, decaying curves of correct solutions. These "zones of confusion" confirm that high uncertainty is a strong proxy for reasoning difficulty.

\textbf{2. The Evolution of Difficulty During Training.} Second, we analyze how these zones of confusion evolve. We compare the uncertainty curves from an early training phase (\textbf{solid lines}) and a late training phase (\textbf{dashed lines}). For incorrect samples, the peaks of uncertainty are more pronounced in the early-to-mid trajectory during early training. As training progresses, the model masters these initial steps, and the uncertainty peaks shift towards later, more complex parts of the reasoning process.

This dual dynamic provides strong, data-driven motivation for our work. It shows that the importance of different steps is highly non-uniform and changes over time. In practice, this means that under uniform subsampling, most of the gradient budget is spent on low-uncertainty, low-instability steps, while only a small number of high-leverage ``zones of confusion'' actually determine success or failure. Any fixed subsampling strategy is therefore inherently suboptimal. This analysis naturally suggests that an adaptive approach is necessary.

\section{Background and Related Work}
\label{sec:background}

\subsection{Diffusion Large Language Models}
dLLMs generate text through an iterative denoising process. Starting from a sequence of fully masked tokens, the model, $f_\theta$, is trained to predict the original text by progressively filling in the masks based on the surrounding context. This non-autoregressive nature allows for parallel decoding and flexible generation orders, distinguishing dLLMs from traditional AR models \cite{nie2025large, ye2025dream}. The core training objective typically minimizes a variant of the negative evidence lower bound (ELBO), training the model to predict original tokens from a corrupted version \cite{he2025mdpo}. This paradigm has been successfully scaled \cite{song2025seed}, extended to long contexts \cite{he2025ultrallada, liu2025longllada}, and adapted for multimodal applications \cite{you2025llada, li2025lavida, yang2025mmada, ma2025consolidating}.

\subsection{Reinforcement Learning for Diffusion Language Models}
Applying policy gradient methods to dLLMs is challenging due to their non-causal structure and the intractability of the marginal likelihood. A series of works has gradually made this feasible. Early trajectory-consistent methods such as diffu-GRPO \cite{zhao2025d1}, MDPO \cite{he2025mdpo}, CT-GRPO \cite{yang2025taming}, and TraceRL \cite{wang2025revolutionizing} bridge the training-inference divide by optimizing the policy over the same multi-step path used at inference time. Complementary approaches like GDPO \cite{rojas2025improving}, LLaDA 1.5 \cite{zhu2025llada1_5}, SPG \cite{wang2025spg}, AWM \cite{xue2025advantage}, and wd1 \cite{tang2025wd1} focus on stabilizing objectives and reducing gradient variance.

Another line of work explores richer process signals and test-time techniques. SAPO \cite{xie2025step}, MDPO \cite{he2025mdpo}, and IGPO \cite{zhao2025inpainting} exploit intermediate denoising steps through step-aware rewards, over-denoising, or inpainting-based hints, while methods such as RFG \cite{chen2025rfg}, Fast-dLLM \cite{wu2025fast, wu2025fastv2}, dllm-cache \cite{liu2025dllm}, Sparse-dLLM \cite{song2025sparse}, CreditDecoding \cite{wang2025creditdecoding}, and dInfer \cite{ma2025dinfer} improve inference-time behavior without changing the training objective.

\textbf{Our Work in Context.} Despite their diversity, these methods share a common design choice: they either treat all denoising steps as equally important, explicitly via uniform subsampling (as in d2-StepMerge \cite{wang2025d2} and related schemes \cite{yang2025taming}) or implicitly by averaging losses uniformly over steps. Our work is the first to directly challenge this assumption by systematically analyzing the non-uniform, dynamic structure of denoising trajectories and then adapting the step selection strategy accordingly. In this sense, ATPO is orthogonal and complementary: it can be layered on top of existing objectives, rewards, and test-time techniques while focusing specifically on \emph{where} along the trajectory to allocate the gradient budget. Rather than redefining the reward or introducing a new objective, ATPO can be plugged into CT-GRPO-style trajectory-aligned scoring, GDPO-like group objectives, or stabilized estimators such as SPG and AWM as a generic, time-step allocation module.

\section{Method: An Adaptive Approach}
\label{sec:method}

Motivated by the non-uniform, dynamic difficulty structure revealed in Section~\ref{sec:motivation} and the limitations of uniform step allocation in existing RL methods summarized in Section~\ref{sec:background}, we propose a simple and intuitive modification to trajectory-based RL: instead of sampling steps uniformly, we should adaptively select them based on their importance. We call this framework \textbf{ATPO}. We refer to any rule that maps a full trajectory together with its step-level difficulty signals to a subset of evaluation steps as an \emph{Adaptive Step Selection Policy}. In our experiments, ATPO instantiates three such policies: RoEC-only Step Selection, CM-Only Step Selection, and our default Hybrid RoEC+CM Step Selection.

\begin{figure}[H]
    \centering
    \includegraphics[width=0.95\columnwidth]{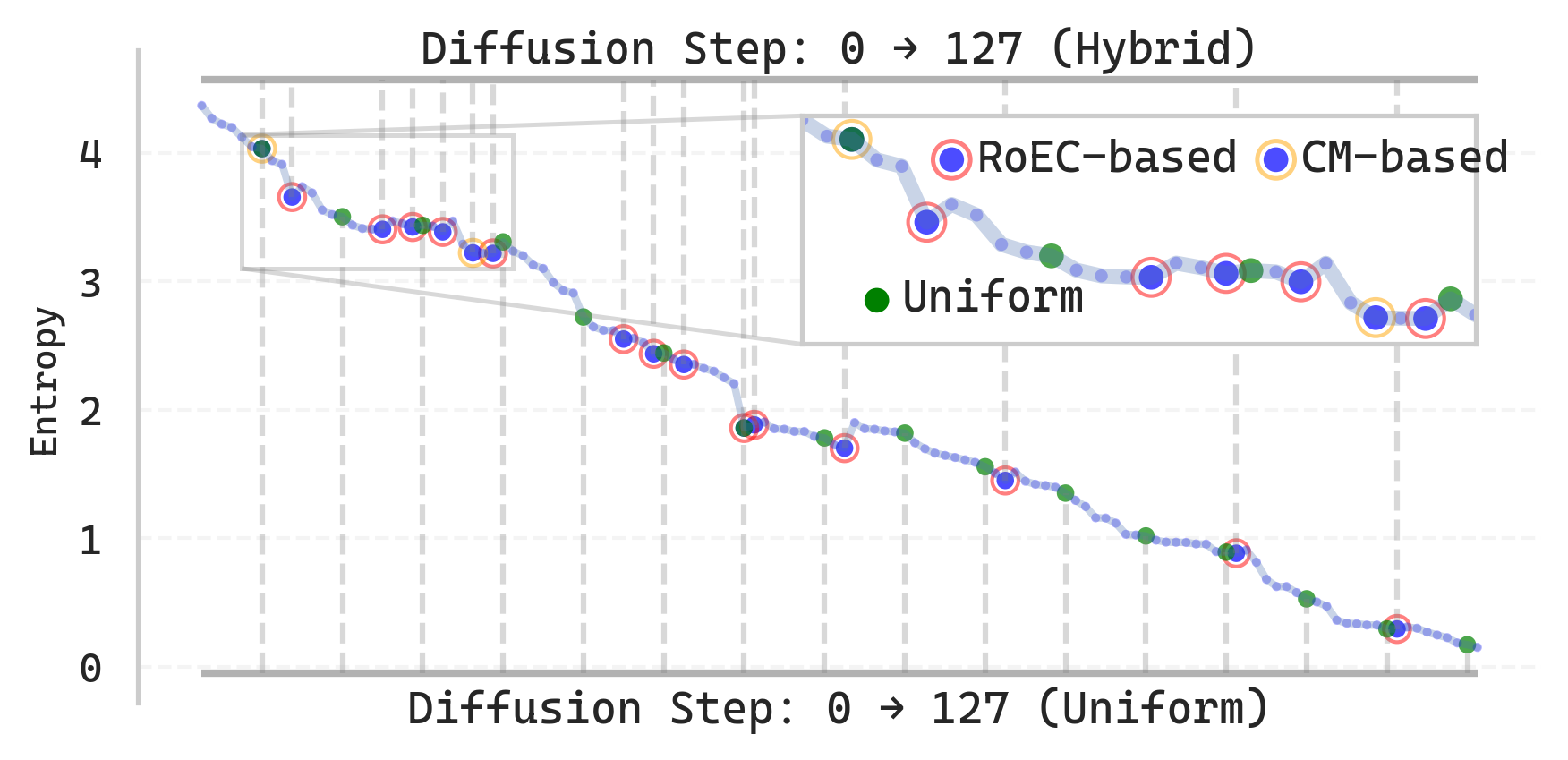}
    \caption{Comparison of entropy curves under different step selection strategies. Uniform step selection (green) results in a smooth but potentially less informative trajectory. Strategies based on Entropy-based Uncertainty and CM are more sensitive to the sharp changes in the early-to-mid diffusion steps, allowing the model to focus on these critical phases.}
    \label{fig:method_segmentation}
\end{figure}

\subsection{Hybrid Adaptive Segmentation}
The core of ATPO is its strategy for dynamically selecting steps. The goal is to focus computation on ``critical junctures'' where the model's decisions are most impactful, which our analysis identified as points of high uncertainty or high instability. This contrasts with prior methods such as d2 \cite{wang2025d2}, whose StepMerge procedure partitions trajectories uniformly and therefore dilutes gradient budget on low-signal regions.

Figure \ref{fig:method_segmentation} highlights that uniform sampling fails to capture the sharp, transient spikes exhibited by diffusion dynamics. Consequently, ATPO treats segmentation as a decision problem driven by two complementary difficulty signals. To enhance computational efficiency and stability, our segmentation strategy operates on batch-averaged dynamics. We compute step-wise entropy and confidence margins for all samples in a batch, average them to yield batch-level mean entropy and confidence-margin curves, and then use these curves for segmentation, ensuring a uniform split scheme across the batch.

\textbf{Hybrid RoEC+CM Step Selection:} Given a target of $N$ segments (requiring $N-1$ split points), we apply a two-stage policy based on the batch-averaged curves. Let $\{H_t\}_{t=1}^T$ and $\{\text{CM}_t\}_{t=1}^T$ denote the batch-averaged entropy and inverse confidence-margin trajectories, respectively, and define the batch-level RoEC curve as
\begin{equation}
    \text{RoEC}_t = 
    \begin{cases}
        0, & t = 1,\\
        |H_t - H_{t-1}|, & t = 2,\dots,T.
    \end{cases}
\end{equation}
We then construct a split set $\mathcal{S}$ of cardinality at most $N-1$ as follows:
\begin{enumerate}
    \item \textbf{Prioritize dynamic instability.} Compute the mean and standard deviation of the RoEC curve, denoted by $\mu_{\text{RoEC}}$ and $\sigma_{\text{RoEC}}$. We first form the RoEC candidate set
    \[
        \mathcal{C}_{\text{RoEC}} = \{\,t \mid \text{RoEC}_t > \mu_{\text{RoEC}} + \sigma_{\text{RoEC}} \,\},
    \]
    and add candidates to $\mathcal{S}$ in increasing order of $t$ until either all elements of $\mathcal{C}_{\text{RoEC}}$ are used or $|\mathcal{S}| = N-1$.
    \item \textbf{Backfill with static uncertainty.} If $|\mathcal{S}| < N-1$, we fill the remaining slots using the static uncertainty curve. We mask out already selected indices in the CM trajectory and choose the remaining time indices with the largest values of $\text{CM}_t$ (highest average inverse confidence margin) to add to $\mathcal{S}$ until $|\mathcal{S}| = N-1$ or no indices remain.
\end{enumerate}
Finally, we deduplicate and sort $\mathcal{S}$, augment it with boundary points $\{0, T\}$, and convert the resulting ordered set $\{0 = b_1 < b_2 < \dots < b_{N+1} = T\}$ into disjoint intervals $\{[b_i, b_{i+1})\}_{i=1}^{N}$. When traces are unavailable or the above heuristics become ill-conditioned (e.g., $T$ is very small or all RoEC values are nearly constant), the procedure falls back to an even partition of the trajectory into $N$ segments, guaranteeing that ATPO never performs worse than the uniform baseline. We note that the threshold $\mu_{\text{RoEC}} + \sigma_{\text{RoEC}}$ was found to be a robust heuristic across our preliminary experiments, but we leave a detailed sensitivity analysis of this hyperparameter to future work.

\subsection{The ATPO Algorithm}
The complete ATPO algorithm is summarized in Algorithm \ref{alg:atpo}. Relative to a standard trajectory-consistent RL loop, ATPO inserts an adaptive instrumentation-and-segmentation module between trajectory generation and policy optimization. During masked diffusion, each step records four synchronized traces—intermediate sequences, token-transfer masks, averaged entropy, and averaged confidence margins—so that the subsequent segmentation operates on sufficient statistics rather than raw tokens. The batch-averaged entropy and confidence-margin curves feed the Hybrid RoEC+CM policy described above, yielding a shared set of split points $\mathcal{S}$ per mini-batch, while the stacked transfer masks preserve which completion tokens were actually edited inside every interval.

\begin{algorithm}[!t]
    \caption{ATPO: Adaptive Trajectory Policy Optimization}
    \label{alg:atpo}
    \begin{algorithmic}[1]
        \STATE \textbf{Input:} dataset $\mathcal{D}$, policy $\pi_\theta$, reference policy $\pi_{\text{ref}}$, reward function $r$, trajectory length $T$, target segments $N$, learning rate $\eta$
        \STATE \textbf{Output:} updated policy parameters $\theta$
        \FOR{each training iteration}
            \STATE Sample a batch of prompts $\mathcal{B}=\{q_j\}$ from $\mathcal{D}$
            \STATE Generate trajectories and traces for all $q_j \in \mathcal{B}$ with $\pi_{\theta_{\text{old}}}$
            \STATE Compute batch-averaged curves $\bar{H}_{1:T}, \overline{\text{CM}}_{1:T}, \overline{\text{RoEC}}_{2:T}$
            \STATE $\mathcal{S} \leftarrow \text{StepSelect}(\bar{H}_{1:T}, \overline{\text{CM}}_{1:T}, \overline{\text{RoEC}}_{2:T}, N)$ \COMMENT{Shared splits}
            \FOR{each trajectory $j$ in the batch}
                \STATE Compute reward $R(y_j)$ and advantage $A(y_j)$ for trajectory $j$
                \STATE $\mathcal{L}_j \leftarrow \text{PPO\_Objective}(\pi_\theta, \pi_{\text{ref}}, \mathcal{S}, A(y_j))$
            \ENDFOR
            \STATE $\mathcal{L} \leftarrow \frac{1}{|\mathcal{B}|} \sum_j \mathcal{L}_j$
            \STATE $\theta \leftarrow \theta + \eta \nabla_\theta \mathcal{L}$
        \ENDFOR
    \end{algorithmic}
    \end{algorithm}

Once $\mathcal{S}$ is fixed, ATPO reuses the original StepMerge estimator to evaluate likelihoods only on the tokens touched within each interval. For a segment $[t_{\text{start}}, t_{\text{end}})$, StepMerge extracts the final state $x_{t_{\text{end}}}$, re-masks the completion positions that changed in this segment, and feeds the masked sequence back through either the current policy or the frozen reference policy. The resulting conditional log-probabilities overwrite only the coordinates selected by the adaptive policy, so the PPO loss, KL penalty, and reward-normalized advantages are all computed on high-leverage steps while keeping the denominator of the log-ratio identical to the uniform baseline. This design makes ATPO strictly orthogonal to choices of reward model, objective variant (e.g., GRPO, PPO-Clip), or architectural accelerations, because those components consume the same per-token log-probabilities regardless of how they were obtained.

\section{Experiments}
\label{sec:experiments}

\definecolor{bestblue}{HTML}{8DB4E2} 

\begin{table*}[!t]
    \centering
    \caption{Performance comparison on reasoning benchmarks. The columns labeled 64, 128, and 256 refer to the generation length.}
    \label{tab:main_results}
    \setlength{\tabcolsep}{8pt}
    \renewcommand{\arraystretch}{1.2}

    \begin{tabular}{lccc ccc ccc ccc}
    \toprule
    \multirow{3}{*}{Model} 
        & \multicolumn{3}{c}{GSM8k} 
        & \multicolumn{3}{c}{Sudoku}
        & \multicolumn{3}{c}{Math500}
        & \multicolumn{3}{c}{Countdown} \\
    \cmidrule(lr){2-4}
    \cmidrule(lr){5-7}
    \cmidrule(lr){8-10}
    \cmidrule(lr){11-13}
        & 64 & 128 & 256 & 64 & 128 & 256 & 64 & 128 & 256 & 64 & 128 & 256 \\
    \midrule


    LLaDA 
        & 68.70 & 76.70 & 78.20 
        & 11.70 & 6.70 & 5.50
        & 26.00 & 32.40 & 36.20        & 20.70 & 19.50 & 16.00 \\

    diffu-GRPO (d1) 
        & \cellcolor{bestblue!40}72.60 
        & \cellcolor{bestblue!40}79.80 
        & \cellcolor{bestblue!40}81.90
        & 18.40 & 12.90 & 11.00
        &  \cellcolor{bestblue!40}33.20 & \cellcolor{bestblue!70}37.20 & \cellcolor{bestblue!40}39.20
        & 33.20 & 31.30 & 37.10 \\

    CJ-GRPO
        & 67.10 
        & 77.75 
        & --
        & \cellcolor{bestblue!70}85.69 
        & \cellcolor{bestblue!70}85.37 
        & --
        & 23.20 & 28.40 & --
        & \cellcolor{bestblue!70}70.80 
        & \cellcolor{bestblue!70}75.59 
        & -- \\

    GDPO
        & 75.06 
        & 81.20 
        & 82.26
        & 14.99 & 24.17 & 25.10
        & 31.40 & 38.00 & 38.20
        & 42.97 & 67.19 & 66.41 \\

    \textbf{ATPO}
        & \cellcolor{bestblue!70}75.51 
        & \cellcolor{bestblue!70}82.49 
        & \cellcolor{bestblue!70}82.94
        & \cellcolor{bestblue!40}72.9 & \cellcolor{bestblue!40}71.48 & \cellcolor{bestblue!40}37.40
        & \cellcolor{bestblue!70}34.00 & \cellcolor{bestblue!40}35.60 & \cellcolor{bestblue!70}39.80
        & \cellcolor{bestblue!40}44.53 & \cellcolor{bestblue!40}70.31 & \cellcolor{bestblue!40}73.69 \\

    d2
        &  & 85.00 & 
        &  & 91.20 & 
        &  & 41.60 & 
        &  & 56.60 &  \\

    \bottomrule
    \end{tabular}

    \vspace{2mm}
    {\footnotesize
    \textbf{Notes.}
    (1) All results are collected directly from the original papers without reproduction.
    (2) Whenever possible, we report configurations that apply RL directly on the base model (LLaDA-8B-Instruct) without an explicit SFT stage; when multiple settings are available (e.g., \cite{zhao2025d1}), we preferentially use the ``pure RL'' variants (such as diffu-GRPO-only) rather than SFT+RL combinations.
    (3) Missing entries ``---'' indicate that the corresponding work did not report results for that sequence length (e.g., \texttt{CJ-GRPO+EOSER} omitted longer sequences).
    (4) The d2 method does not explicitly specify generation hyperparameters; we report its raw numbers as presented in the paper.
    }
\end{table*}

\subsection{Experimental Setup}

\textbf{Framework and Model.} Our experimental framework is built upon the open-source \texttt{d1} project \cite{zhao2025d1}, ensuring our methodology is transparent and reproducible. We use the LLaDA-8B-Instruct model \cite{nie2025large, zhao2025d1} as our base dLLM for all experiments, and conduct training on a cluster of 8 NVIDIA H800-80G GPUs.

\textbf{Training Details.} To ensure a fair comparison with prior work, we adhere as closely as possible to the training configuration established by \cite{zhao2025d1}. We employ LoRA \cite{hu2022lora} for parameter-efficient fine-tuning, with a rank of $r=128$ and a scaling factor of $\alpha=64$. For reinforcement learning, the generation sequence length is fixed at 256 tokens, with a per-GPU batch size of 6 and 2 gradient accumulation steps, resulting in an effective batch size of 96. We use the AdamW optimizer with a learning rate of $3 \times 10^{-6}$ and gradient clipping at 0.2. To accelerate training, we leverage FlashAttention-2 \cite{dao2022flashattention}. Unless stated otherwise, core components such as the reward model, the PPO-style RL objective, and the underlying data pipeline remain identical across all of our variants. Within this unified framework, our key intervention is the step selection strategy (uniform vs.\ adaptive RoEC+CM), allowing us to isolate the impact of dynamic gradient allocation while keeping model architecture, reward, and optimizer unchanged. The overhead of this adaptive selection is minimal: the analysis pass to compute uncertainty metrics and select steps adds approximately 2-3\% to the wall-clock time per training step compared to the uniform sampling baseline.

\textbf{Benchmarks and Evaluation.} We evaluate all methods on a diverse suite of reasoning benchmarks:
\begin{itemize}
\item \textbf{GSM8K} \cite{cobbe2021training}: Grade school math word problems.
\item \textbf{Math500} \cite{hendrycks2020measuring}: Competition-level mathematics problems.
\item \textbf{Sudoku}: A logic puzzle task requiring constraint satisfaction.
\item \textbf{Countdown}: A number game requiring arithmetic reasoning to reach a target number.
\end{itemize}
For all benchmarks, we report task-specific accuracy using greedy decoding and evaluate across multiple generation lengths (64, 128, and 256) to assess performance under varying computational budgets.

\subsection{Comparison Methods and Main Results}
A central tenet of our work is to enhance the reasoning capabilities of a base dLLM through direct reinforcement learning, without an initial supervised fine-tuning (SFT) stage. This approach aims to directly instill reasoning abilities via RL. Consequently, when comparing to prior methods we focus on configurations that are also reported as operating directly on the LLaDA-8B-Instruct base model without an explicit SFT phase whenever such numbers are available. In particular, we include \textbf{diffu-GRPO (d1)} \cite{zhao2025d1}, i.e., the pure diffu-GRPO variant without SFT that serves as the foundational trajectory-consistent RL recipe; \textbf{GDPO} \cite{rojas2025improving}, which focuses on variance reduction; \textbf{d2} \cite{wang2025d2}, a framework utilizing uniform subsampling (StepMerge); and \textbf{CJ-GRPO} \cite{yang2025taming}, which explores trajectory consistency.

To ensure a broad and objective comparison, the results for these baseline methods are cited directly from their original publications. Whenever the original work reports pure-RL configurations directly on LLaDA-8B-Instruct (e.g., diffu-GRPO-only variants in \cite{zhao2025d1}), we use these numbers as our primary reference points; for methods with additional SFT or architectural variations, we report the configuration that most closely matches our setting. We align inference conditions (e.g., generation length) with the reported settings whenever possible, but do not re-train external baselines, as their full training configurations are not always public, making a perfectly controlled re-implementation infeasible. All ablations and uniform-vs-adaptive comparisons are conducted within our \texttt{d1}-based implementation under shared hyperparameters, so relative improvements among our own variants are directly comparable.

Table \ref{tab:main_results} presents the main experimental results. These experiments validate our thesis that, under a comparable computational budget and RL objective, adapting to trajectory dynamics is more effective than uniform subsampling. ATPO consistently achieves strong performance, particularly on complex tasks and longer sequences where the non-uniform structure of the denoising trajectory is most pronounced. For instance, on GSM8K with a sequence length of 256, ATPO achieves an accuracy of \textbf{82.94}, outperforming all compared direct-RL methods. The robust improvements across diverse tasks like Sudoku and Countdown further underscore that reallocating gradient updates towards high-uncertainty, high-instability regions is a general and effective strategy.

\subsection{Training Stability and Ablation Analysis}

Beyond absolute performance, a key benefit derived from our analysis is improved training stability. To isolate the impact of step selection, we conduct an ablation study under a fixed RL setup where we keep the model, reward, and objective unchanged, and only vary the Adaptive Step Selection Policy. To ensure a fair comparison, all variants were trained with identical hyperparameters. Specifically, we compare four variants: \emph{uniform} subsampling (the standard choice in prior work), \emph{ATPO with RoEC-only Step Selection}, \emph{ATPO with CM-Only Step Selection}, and our full \emph{ATPO with Hybrid RoEC+CM Step Selection}.

\begin{figure}[H]
\centering\includegraphics[width=0.8\columnwidth]{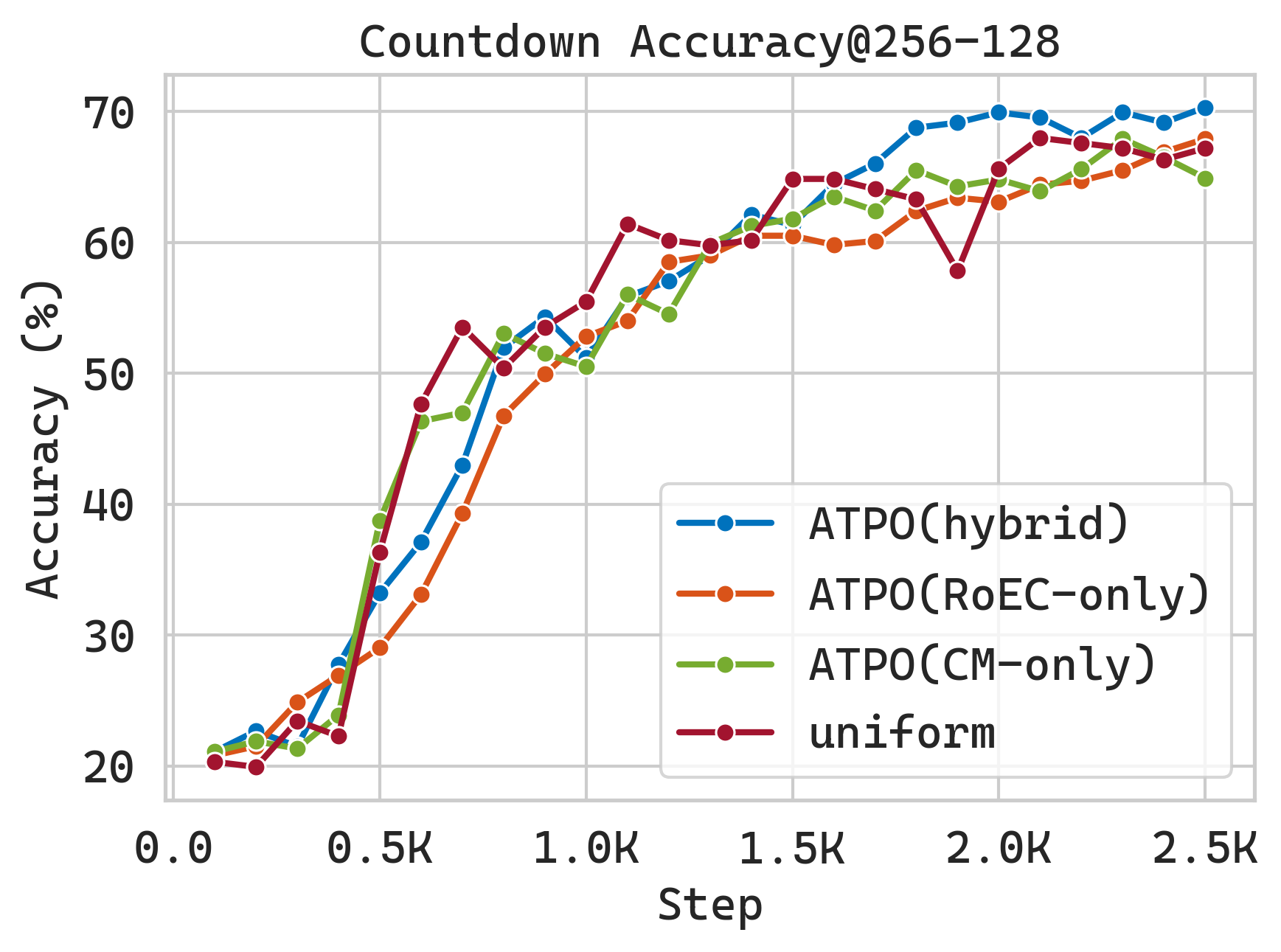}
\caption{Checkpoint accuracy over training steps under four step selection strategies: uniform subsampling, ATPO with RoEC-only Step Selection, ATPO with CM-Only Step Selection, and ATPO with Hybrid RoEC+CM Step Selection. The hybrid policy yields both the highest final accuracy and the smoothest training curve.}
\label{fig:training_curves}
\end{figure}

Figure \ref{fig:training_curves} plots test accuracy as a function of training steps on a representative reasoning benchmark (Math500 with sequence length 256) for these four variants. We observe three consistent patterns. First, \textbf{ATPO (hybrid)} dominates throughout training: after an initial warm-up, it steadily climbs past 60\% and reaches 70.31\% at step 2500, with only minor fluctuations, indicating both faster convergence and a higher final plateau. Second, \textbf{ATPO (RoEC-only)} yields smoother curves than uniform subsampling but is noticeably \emph{slower} in the mid-training regime: from steps 500 to 1800 it consistently lags behind uniform, only catching up and slightly surpassing it near the end (67.9\% vs.\ 67.19\%), and still remains several points below the hybrid variant. This aligns with our hypothesis that RoEC-only selection over-focuses on persistently high-entropy but low-impact regions. Third, \textbf{ATPO (CM-only)} behaves as a more aggressive yet brittle strategy: it occasionally spikes above uniform in the mid-training phase, but exhibits larger oscillations and eventually underperforms both uniform and RoEC-only (ending at 64.9\%), confirming that purely local token-level conflicts are not a reliable proxy for global reasoning progress.

Overall, these results support our analysis-driven design: RoEC and CM capture complementary aspects of trajectory difficulty, and only their hybrid combination reliably concentrates gradient updates on the few high-leverage steps that determine success or failure, yielding both higher accuracy and markedly more stable training dynamics.

\section{Discussion}
\label{sec:discussion}

This paper's primary contribution is an analysis of the reasoning dynamics within dLLM trajectories. We moved beyond the simplifying assumption of uniform step importance and showed, both formally and empirically, that these pathways are characterized by non-uniform, dynamic uncertainty. The key insight is that the model's "struggle"—quantified by metrics like RoEC and CM—is not noise to be averaged out, but a signal to be leveraged.

Our adaptive method, ATPO, was proposed as a direct consequence and validation of this analysis. Its strong performance is evidence that an analysis-driven approach to RL algorithm design is effective. ATPO synthesizes goals from prior work: it maintains \textbf{trajectory consistency} (like CT-GRPO), addresses \textbf{computational efficiency} (like d2), and improves \textbf{training stability}, all through the single, unifying principle of adapting to trajectory dynamics.

It is also instructive to contrast ATPO with several closely related lines of work. SAPO \cite{xie2025step} focuses on designing step-aware \emph{rewards}, assuming that the locations where these rewards are applied are already given, while IGPO \cite{zhao2025inpainting} targets early-stage exploration by injecting inpainting hints, and MDPO \cite{he2025mdpo} introduces over-denoising-based \emph{sample} filtering rather than time-step allocation. RFG \cite{chen2025rfg} and other test-time scaling methods operate purely at inference time, modifying the decoding process without changing the training dynamics. In contrast, ATPO is orthogonal to these methods: instead of redefining \emph{what} to optimize (objectives, rewards, or data), it focuses on \emph{where} along the trajectory to allocate the limited gradient budget. In practice, the adaptive step-selection module can be plugged into CT-GRPO-style trajectory-aligned scoring (losses computed only over newly unmasked tokens), GDPO-like log-mean estimators and other stabilized objectives such as SPG or AWM, while leaving the underlying RL objective, reward function, and data pipeline unchanged; by simply toggling between adaptive and uniform step selection, ATPO can recover behavior similar to d2 \cite{wang2025d2}, underscoring its role as a lightweight and generally compatible component.

However, our analysis is not exhaustive. The proposed metrics (Entropy, CM, RoEC) are effective heuristics, but other, potentially more powerful, measures of trajectory instability may exist. The overhead of the "analysis pass" to compute these metrics is a practical trade-off, although our results suggest it is a worthwhile one. Finally, this work focuses on reasoning tasks; how these dynamics manifest in other domains, like creative writing or dialogue, remains an open and interesting question for future research.

\section{Conclusion}
\label{sec:conclusion}

This paper presented an in-depth analysis of the denoising trajectory dynamics in Diffusion Language Models. We challenged the prevailing assumption of uniformity in trajectory-based reinforcement learning and demonstrated that the reasoning process is characterized by critical, non-uniform, and dynamic moments of uncertainty. By formalizing and empirically identifying these "surprise points" using information-theoretic metrics, we revealed the limitations of fixed, uniform subsampling strategies.

As a direct consequence of our analysis, we introduced ATPO, a policy optimization framework that replaces uniform subsampling with an adaptive step selection strategy. By intelligently focusing computation on the highest-leverage moments in the reasoning chain, ATPO achieves significant improvements in performance and training stability. Conceptually, our work adds an orthogonal dimension to the existing literature: instead of proposing yet another objective, reward, or estimator, we study and optimize \emph{where} along the trajectory the gradient budget should be allocated. This work validates that a deeper, analysis-driven understanding of the underlying generative process---and, in particular, of the dynamic structure of generative trajectories---is a promising direction for developing more intelligent and efficient reinforcement learning algorithms for next-generation language models.

\section*{Acknowledgment}

\end{document}